\documentclass{article}
\usepackage{PRIMEarxiv}

\pdfoutput=1
\usepackage[utf8]{inputenc}
\usepackage[T1]{fontenc}
\usepackage{lmodern}
\usepackage{microtype}
\usepackage{url}
\usepackage{hyperref}

\usepackage{amsmath,amssymb,amsfonts,mathrsfs,mathtools,amsthm,bm,bbm}

\usepackage{booktabs}
\usepackage{multirow}
\usepackage{array}
\usepackage[table]{xcolor}
\usepackage{graphicx}
\graphicspath{{image/}}
\usepackage{caption}
\usepackage{subcaption}
\usepackage{adjustbox}
\usepackage{float}
\usepackage{tikz}
\usepackage{nicefrac}

\usepackage{algorithm}
\usepackage{algorithmicx}
\usepackage{algpseudocode}

\usepackage{xcolor}
\usepackage{enumitem}
\usepackage{listings}
\usepackage{pifont}
\usepackage{comment}

\theoremstyle{plain}

\theoremstyle{remark}

\theoremstyle{definition}

\usepackage{amsmath,amsfonts,bm}

\def\eqref#1{equation~\ref{#1}}

\def\1{\bm{1}}

\DeclareMathAlphabet{\mathsfit}{\encodingdefault}{\sfdefault}{m}{sl}
\SetMathAlphabet{\mathsfit}{bold}{\encodingdefault}{\sfdefault}{bx}{n}

\hypersetup{
    colorlinks=true,
    linkcolor=blue,
    citecolor=blue,
    urlcolor=blue
}

\title{No Pose Estimation? No Problem: Pose-Agnostic and Instance-Aware Test-Time Adaptation for Monocular Depth Estimation}

\author{
    Mingyu~Sung,
    Hyeonmin~Choe,
    Il-Min~Kim, 
    Sangseok~Yun, 
    and~Jae-Mo~Kang%
    \thanks{
        Mingyu Sung, Hyeonmin~Choe and Jae-Mo Kang are with the Department of Artificial Intelligence, Kyungpook National University, Daegu, South Korea (Corresponding author: Jae-Mo Kang, e-mail: jmkang@knu.ac.kr).
    }%
    \thanks{
        Il-Min Kim is with the Department of Electrical and Computer Engineering, Queen's University, Kingston, K7L 3N6, Canada.
    }%
    \thanks{
        Sangseok Yun is with the Department of Information and Communications Engineering, Pukyong National University, Busan 48513, South Korea (Corresponding author: Sangseok Yun, e-mail: ssyun@pknu.ac.kr).
    }
}

\begin{document}
\maketitle

\begin{abstract}
Monocular depth estimation (MDE), inferring pixel-level depths in single RGB images from a monocular camera, plays a crucial and pivotal role in a variety of AI applications demanding a three-dimensional (3D) topographical scene. 
In the real-world scenarios, MDE models often need to be deployed in environments with different conditions from those for training.
Test-time (domain) adaptation (TTA) is one of the compelling and practical approaches to address the issue.
Although there have been notable advancements in TTA for MDE, particularly in a self-supervised manner, existing methods are still ineffective and problematic when applied to diverse and dynamic environments. 
To break through this challenge, we propose a novel and high-performing TTA framework for MDE, named PITTA. Our approach incorporates two key innovative strategies: (i) pose-agnostic TTA paradigm for MDE and (ii) instance-aware image masking.
Specifically, PITTA enables highly effective TTA on a pretrained MDE network in a pose-agnostic manner without resorting to any camera pose information.
Besides, our instance-aware masking strategy extracts instance-wise masks for dynamic objects (e.g., vehicles, pedestrians, etc.) from a segmentation mask produced by a pretrained panoptic segmentation network, by removing static objects including background components. These masks serve as informative and useful cues for MDE during TTA and are used to selectively mask the depth map (i.e., output of the MDE network).
To further boost performance, we also present a simple yet effective edge extraction methodology for the input image (i.e., a single monocular image) and depth map.
Based upon these strategies, we develop a powerful TTA strategy for the MDE network by introducing and balancing two customized loss functions, namely, depth-refining loss and edge-guided loss.
Extensive experimental evaluations on DrivingStereo and Waymo datasets with varying environmental conditions demonstrate that our proposed framework, PITTA, surpasses the existing state-of-the-art techniques with remarkable performance improvements in MDE during TTA.
\end{abstract}

\section{Introduction}
Monocular depth estimation (MDE) is a computer vision task that predicts the depth of each pixel in a single RGB image from a monocular camera {\color{black}\cite{rajpal2023high}}.\footnote{
Such an image will be referred to simply as a single monocular image throughout this paper whenever there is no ambiguity. 
} 
As a key technology for three-dimensional (3D) perception, MDE is envisioned to play a paramount and essential role in numerous AI applications such as autonomous driving, robotics, augmented reality, scene understanding, etc. 
In these applications, MDE models or networks are often required to be deployed in dynamic environments under diverse conditions---even rather different from those during training---where distribution or domain of real data varies continually {\color{black}\cite{li2023test, wang2022continual}}, e.g., due to variations in lighting, weather, objects, etc.
Test-time adaptation (TTA) is a compelling and effective solution to cope with this issue by enabling MDE networks to adapt to new, unseen domains or environments during inference with neither retraining nor access to source datasets used for training {\color{black}\cite{kuznietsov2021comoda, li2023test}}. 

\begin{figure*}[t]
    \centering
    \vspace{-2em}
    \begin{minipage}{0.44\textwidth} 
        \centering
        \centerline{\includegraphics[width=\linewidth]{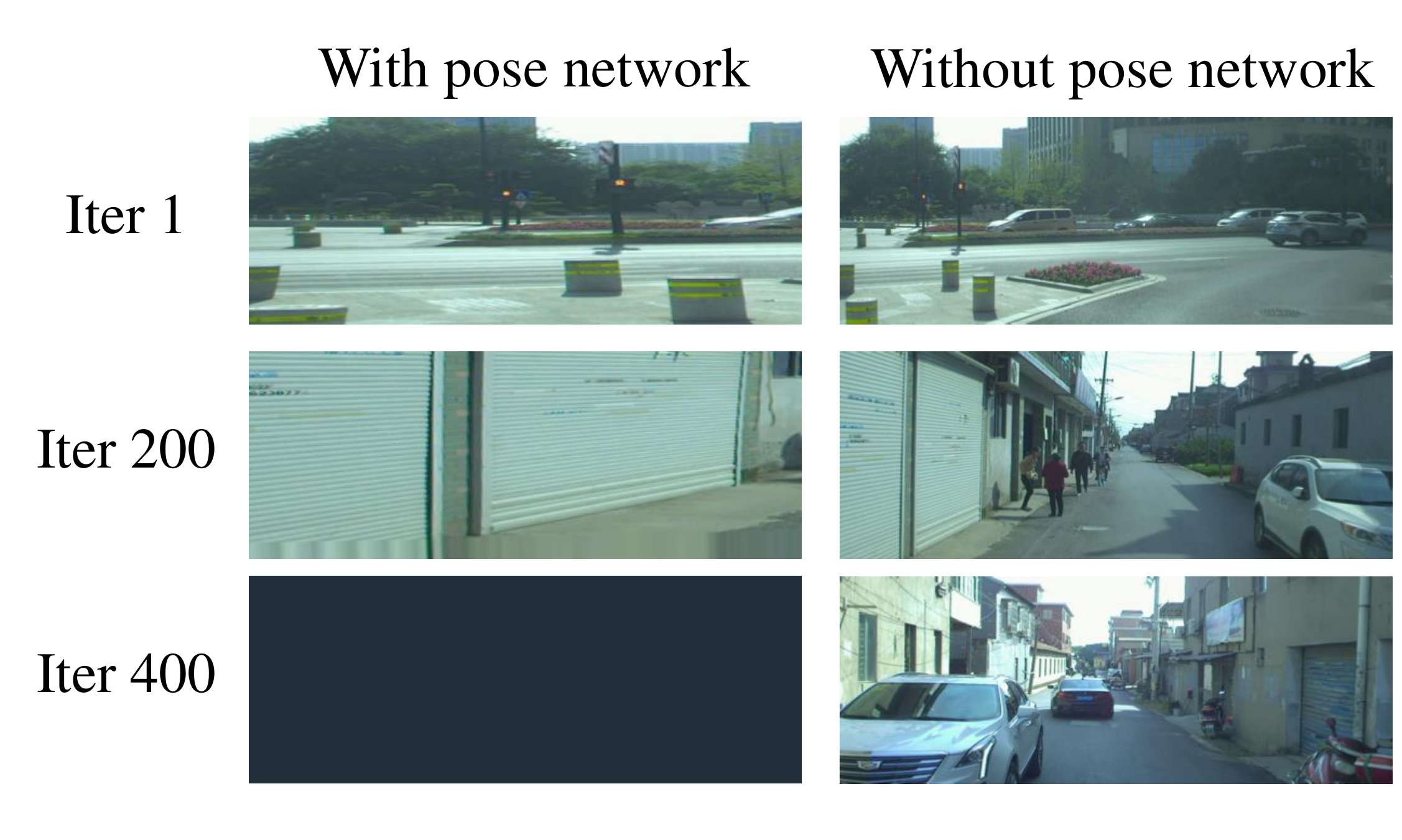}}    
    \end{minipage}\hfill 
    \begin{minipage}{0.56\textwidth} 
        \centering
        \centerline{\includegraphics[width=1.03\linewidth]{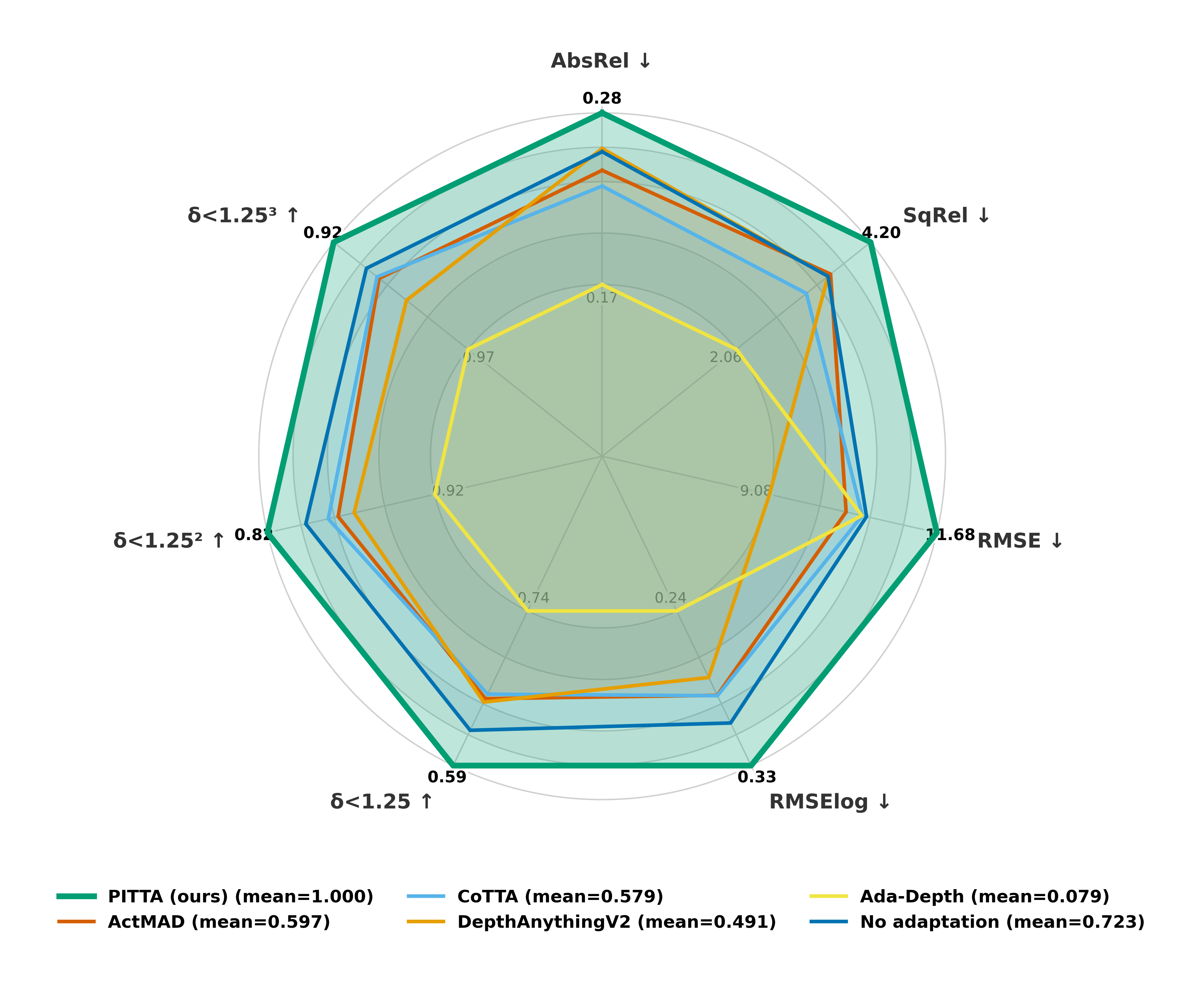}}
    \end{minipage}
    \vspace{-1.5em}
    \caption{\small (a) Reconstructed images during TTA based on SfM assumption by following \cite{godard2019digging} with and without pose estimation network. 
    (b) TTA performance of our proposed method and other competing methods over 7 MDE metrics averaged over 4 different test cases on DrivingStereo and Waymo datasets.}
    \label{fig:intro}
    \vspace{-1em}
\end{figure*}

According to these demands, several TTA methods for MDE have been developed in the recent studies, such as {\color{black}CoMoDa \cite{kuznietsov2021comoda} and Ada-Depth \cite{li2023test}}.
Other TTA approaches applicable to MDE scenarios have also been reported in the literature, such as {\color{black}ActMAD \cite{mirza2023actmad} and CoTTA \cite{wang2022continual}}.
Despite merits and efficacy of these existing methods, several major technical limitations still remain.
One critical and practical concern is that the MDE approaches in {\color{black} \cite{kuznietsov2021comoda, li2023test}} are resorting to the Structure-from-Motion (SfM) assumption {\color{black}\cite{zhou2017unsupervised}}, which is however is highly likely to be violated in dynamic environments with domain shifts---realistic TTA situations.
In addition to this, due to the reliance on the SfM assumption, adapting MDE networks using the TTA methods in {\color{black}\cite{kuznietsov2021comoda, li2023test}} should be accompanied by additional adaptation of a separate network called pose estimation network. 
Thus, the performance of adapting the MDE network is sensitive and prone to the adaptability of the pose estimation network, but proper or optimal adaptation of the pose estimation network in practical TTA setups is still nontrivial and difficult, to our knowledge.
Indeed, adapting the pose estimation network by directly extending a naive strategy based on the SfM assumption could lead to improper adaptations of both MDE network and pose estimation network during TTA, even exhibiting performance inferior to that of no adaptation.
This serious phenomenon is empirically demonstrated in Fig. \ref{fig:intro}(a) with an illustrative example and in more detail in Appendix D.1 with visualization results.
Besides, the performance of other TTA approaches in {\color{black}\cite{mirza2023actmad, wang2022continual}} is generally suboptimal as they are not specialized (and thus, not suited) for MDE tasks.
Furthermore, in the approaches of {\color{black}\cite{li2023test, mirza2023actmad}}, it is implicitly assumed that metadata and statistics of source datasets are available, respectively, which however may not be valid in strict real-world TTA settings without any access to source datasets. 
Above all, the most significant limitation is that the existing approaches in {\color{black}\cite{kuznietsov2021comoda, li2023test, mirza2023actmad, wang2022continual}} do not fully utilize available information for MDE during TTA, thereby hindering the potential and possibility for further performance improvements.

To break through these limitations, we introduce a pose-agnostic, instance-aware TTA framework for MDE, named PITTA, which achieves notable performance improvements as demonstrated in Fig. \ref{fig:intro}(b). 
One of the key technical innovations in our framework is that we present a novel pose-agnostic TTA paradigm for MDE, which does not require the SfM assumption. Unlike the existing approaches in {\color{black}\cite{kuznietsov2021comoda, li2023test}}, our pose-agnostic TTA paradigm enables a pretrained MDE network to be adapted with no camera pose information at all, eliminating the need for the adaptation of the pose estimation network during the adaptation of the MDE network.
Another technical innovation is that we also devise a novel instance-aware masking strategy, which substantially enhances the adaptability of the MDE network during TTA. 
In this strategy, useful and valuable information on dynamic object instances (including vehicles, pedestrians, etc.) in monocular images are effectively utilized for MDE during TTA by means of so-called instance-wise masks, rather than using camera pose information as in prior works {\color{black}\cite{kuznietsov2021comoda, li2023test}}.
On top of these, our framework further improves the performance by utilizing detected object boundaries as additional information for MDE and by balancing it with the dynamic instance information, respectively, through effective edge extraction mechanism and loss function design.
Our key technical contributions and breakthroughs in this work include the followings: 
\begin{itemize}
\item We propose PITTA, a novel pose-agnostic and instance-aware TTA framework for MDE, which enables high-performing and effective adaptation of a pretrained MDE network in a pose-agnostic manner without resorting to any camera pose information.

\item We devise an innovative and effective instance-aware masking strategy that can substantially enhance the adaptability of the MDE network during TTA by exploiting instance-wise masks for dynamic objects. 

\item We present an effective edge extraction mechanism to utilize detected object boundaries as additional informative cues for further enhancing MDE during TTA. 

\item We introduce two customized loss functions, depth-refining loss and edge-guided loss. Balancing these two loss functions via selective update of parameters enables appropriate adaptation of the MDE network.

\item Through extensive experimental validation on widely used datasets, DrivingStereo dataset \cite{yang2019drivingstereo} and Waymo dataset \cite{sun2020scalability}, under diverse environmental conditions, we empirically demonstrate that PITTA markedly surpasses the existing state-of-the-art (SOTA) techniques in various MDE performance metrics and substantially enhances the adaptability of diverse MDE networks.




\end{itemize}

\section{Related Works}
\textbf{Monocular Depth Estimation.} 
In the literature, MDE has been widely and intensively studied as it is a long-standing and challenging task in the areas of computer vision {\color{black}\cite{10313067}}.
MDE techniques can be roughly divided into two categories depending on the availability of labels (i.e., ground-truth depth values) during training---(i) supervised approaches such as {\color{black} Adabins \cite{bhat2021adabins}, multi-scale MDE network \cite{eigen2014depth}, and NewCRFs \cite{yuan2022neural}}; and (ii) self-supervised approaches such as {\color{black}MonoDepth2 \cite{godard2019digging}, SGDepth \cite{ guizilini2020semantically}, HR-Depth \cite{lyu2021hr}, Lite-Mono \cite{zhang2023lite}, and MonoViT \cite{zhao2022monovit}}.
More details of related works on MDE with comprehensive literature review can be found in {\color{black}Appendix \ref{app_B1}}.
The supervised MDE techniques in {\color{black}\cite{bhat2021adabins, eigen2014depth, yuan2022neural}} have enough potential to predict pixel-level depths with higher accuracy than the self-supervised ones.
Nevertheless, a universal and practical concern is that acquiring labeled datasets is generally costly and time-consuming as they should be collected exhaustively in practice using LIDAR, and/or depth-measuring equipment like RGB-D sensors {\color{black}\cite{geiger2012we, 9394752}}.
To address this issue, in the self-supervised MDE approaches {\color{black}\cite{godard2019digging, guizilini2020semantically, lyu2021hr, zhang2023lite, zhao2022monovit}},   
the depth values are estimated without the ground-truth labels, only from monocular images over different frames by invoking the SfM assumption.
Unfortunately, however, the existing MDE methods in {\color{black}\cite{godard2019digging, guizilini2020semantically, lyu2021hr, zhang2023lite, zhao2022monovit}} all assumed that the environments or domains remain unchanged during both training and inference, and thus, these methods may perform poorly in practical TTA scenarios.

\textbf{Test-Time Adaptation for Monocular Depth Estimation.}
Research on TTA for MDE remains largely unexplored in the literature, to our knowledge, and only a few works {\color{black}\cite{kuznietsov2021comoda, li2023test}} have touched the issue in a self-supervised manner. 
More details of these works are provided in {\color{black}Appendix B}.
However, the MDE approaches in {\color{black}\cite{kuznietsov2021comoda, li2023test}} are based on extensions of the earlier self-supervised MDE methods---such as in {\color{black}\cite{godard2019digging, yuan2022neural}}---developed for non-TTA scenarios.
Unfortunately, therefore, these methods are still ineffective and problematic when applied to diverse and dynamic environments, primarily on account of the reliance on the SfM assumption and the difficulty in properly adapting the pose estimation network.

\section{Our Approach: PITTA}
In this work, the MDE task is performed via an off-the-shelf MDE network---{\color{black}such as in \cite{godard2019digging, guizilini2020semantically, lyu2021hr, zhang2023lite, zhao2022monovit}}---that outputs the pixel-wise depth estimates by taking a single monocular image as the input.
Our goal is to adapt (or adjust) the MDE network pretrained on source (or training) datasets of monocular images to target (i.e., another) datasets---being in different domains or exhibiting distributional shifts from the source datasets---during the test (or inference) phase. We consider a realistic online situation where each sample (i.e., a single monocular image) in the target datasets is accessible sequentially, while the samples in the source datasets are not available at all. To accomplish this goal and break through the relevant technical challenges, we propose PITTA, a novel and high-performing TTA framework for MDE.
The overall computation procedure of our framework is presented in Algorithm 1 of Appendix C and detailed in what follows.

\begin{figure*}[t]
\centering
\centerline{\includegraphics[width=1.02\textwidth]{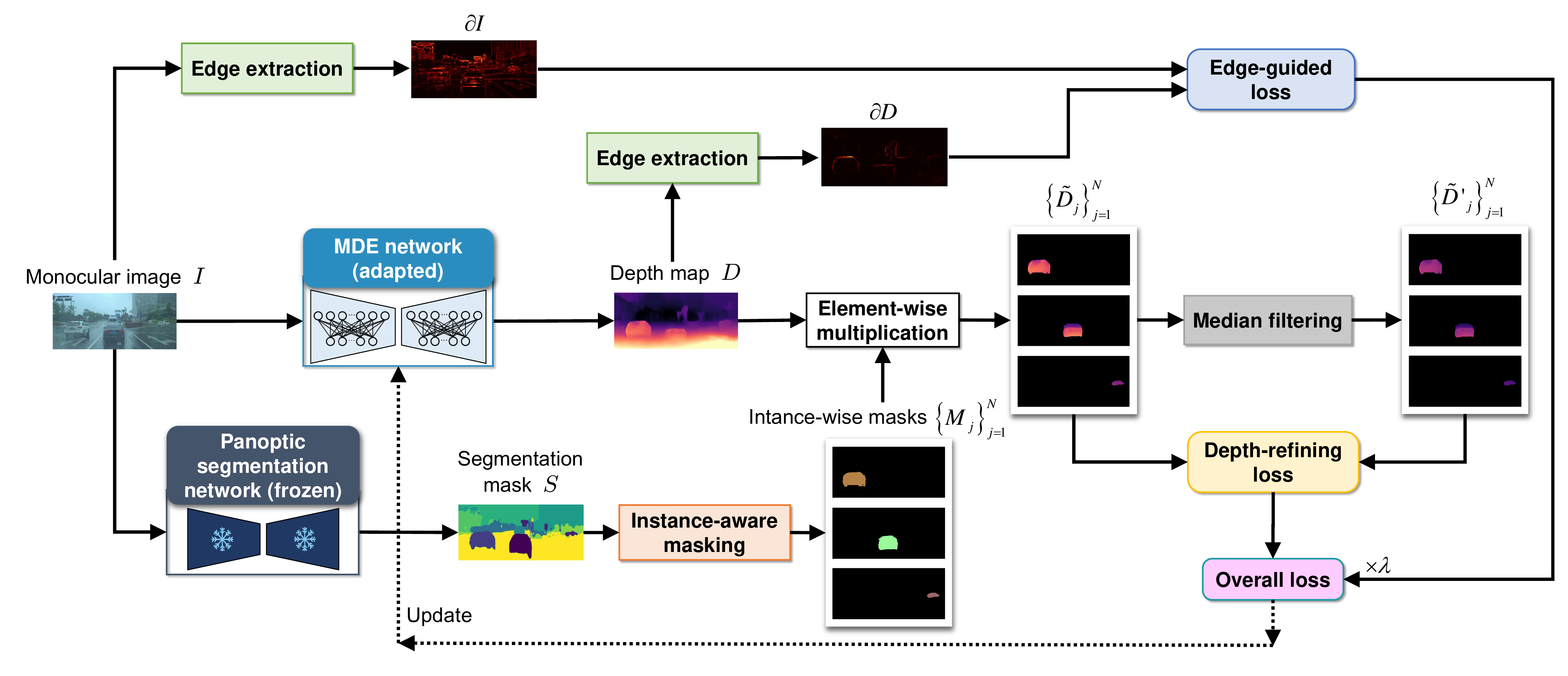}} 
\caption{\small Overall architecture and schematic diagram of our TTA framework to adapt a pretrained MDE network given sequences of single RGB images from a monocular camera. Detailed computation procedures are presented in Algorithm 1 of Appendix C.}
\label{fig:arch}
\vspace{-1em}
\end{figure*}

\subsection{Overall Architecture}

\textbf{Overview.} The overall architecture and schematic diagram of PITTA are depicted in Fig. \ref{fig:arch}.
As can be seen from Fig. \ref{fig:arch}, our framework contains several modules. 
Firstly, there are two types of pretrained off-the-shelf networks: one is a pretrained MDE network (to be adapted) and the other is a pretrained panoptic segmentation network (frozen during the adaptation of the MDE network). 
On top of these networks, an additional innovative design, called instance-aware masking, is further employed in PITTA along with an effective edge extraction strategy for technical advancement and performance improvement of MDE during TTA.
Further details of these modules are provided in the following.


\textbf{MDE Network.} At each time step, the MDE network predicts the depth of each pixel in the input image (i.e., a single RGB image from a monocular camera) denoted by $I \in  \mathbb{R}^{H \times W \times C}$ in Fig. \ref{fig:arch}, where $H$, $W$, and $C$ denote height, width, and channel, respectively. 
The output of the MDE network is a collection of depth estimates for all the pixels, called depth map, denoted by $D \in  \mathbb{R}^{H \times W}$ in Fig. \ref{fig:arch}.
In PITTA, we primarily adopt a pretrained MDE network of MonoDepth2 in \cite{godard2019digging}, which is based on a U-Net architecture using ResNet-18 as the encoder. 
Further details of this network is provided in {\color{black}Appendix D.1.}
Also, in PITTA, we propose to adapt only parts---not the whole part---of the adopted MDE network to the test datasets during TTA. The reasons for this will be detailed later in Section 3.2.
In general, another off-the-shelf MDE network such as in {\color{black}\cite{guizilini2020semantically, lyu2021hr, zhang2023lite, zhao2022monovit}} can also be adopted in our framework.


\textbf{Pose-Agnostic TTA Paradigm.} 
The existing MDE approaches in {\color{black}\cite{godard2019digging, guizilini2020semantically, kuznietsov2021comoda, li2023test, lyu2021hr, zhang2023lite, zhao2022monovit}} relying on the SfM assumption are ineffective and problematic for TTA as they require the knowledge of camera pose information with continual adjustment, which is however rather challenging in practice. 
For more detailed exposition, let $(x,y)$ denote the two-dimensional (2D) coordinate of each pixel of the image image $I$ in a particular frame and $(x',y')$ denote the 2D coordinate of the corresponding pixel in another frame.
Also, let $D(x,y) > 0$ denote the depth value for pixel $(x,y)$, i.e., pixel intensity of $D$ at $(x,y)$. The common approach in the the existing methods is to retrieve the depth information from images in different frames based on the following 3D geometry relationship: 
\begin{align}
\small
\label{eq_1}
    K   \left(   R  K^{-1}  \begin{bmatrix}
        x \\ y  \\ 1
    \end{bmatrix} D(x,y)  +  t  \right)   =  z'  \begin{bmatrix}
        x' \\ y'  \\ 1
    \end{bmatrix}
\end{align}
where $K \in  \mathbb{R}^{3 \times 3}$ is a (known) camera intrinsic matrix and $z'$ corresponds to the last entry of the $3 \times 1$ vector in the left side of \eqref{eq_1}.
Also, $R \in  \mathbb{R}^{3 \times 3}$ and $t \in  \mathbb{R}^{3 \times 1}$ represent rotation matrix and transition vector, respectively, which are typically unknown and predicted by a pretrained pose estimation network such as in {\color{black}\cite{godard2019digging}}. 
Note that to make use of the relationship in \eqref{eq_1} for TTA (i.e., adaptation of $D(x,y)$), it is required to adapt $R$ and $t$ accordingly as well.
To this end, proper adaptation of the pose estimation network is essential, which is however nontrivial and rather challenging.
To overcome this critical challenge effectively and uncomplicatedly, in PITTA, we present a new and efficient TTA paradigm for MDE in a camera pose-agnostic manner.
The core idea of our TTA approach is to adapt the MDE network with no (prior) knowledge of camera pose information:
\begin{align}
\small
    R = \begin{bmatrix}
        1  &  0   &  0  \\   0  &  1  &  0  \\ 0 &  0  &  1
    \end{bmatrix} ,   \quad    t = \begin{bmatrix}
        0 \\ 0  \\ 0
    \end{bmatrix} ,
\end{align}
by simply presuming that there is no pixel transition across images in different frames.
Our TTA approach is indeed useful and effective as it liquidates the need for acquisition of accurate camera pose estimates as well as reliance on a pose estimation network during the adaptation of the MDE network.
Furthermore, it is worth noting that even without the exploitation of camera pose information, our framework markedly surpasses the existing techniques in a variety of TTA tasks for MDE, as will be demonstrated by the extensive and intensive experimental validations presented in Section 4.2.

\textbf{Panoptic Segmentation Network.} In PITTA, we also employ a pretrained panoptic segmentation network, the output of which, called panoptic segmentation mask, denoted by $S \in  \mathbb{R}^{H \times W}$ in Fig. \ref{fig:arch} will be used subsequently for the instance-aware masking of depth map $D$.
We primarily adopt a network in \cite{cheng2022masked}, Mask2Former, which utilizes the tiny version of Swin Transformer \cite{liu2021swin} as backbone.
Further details of Mask2Former are provided in {\color{black}Appendix C.2.}
Another off-the-shelf panoptic segmentation network such as in {\color{black}\cite{hu2023you,kim2020vps}} can also be employed in PITTA.
The panoptic segmentation network produces a panoptic segmentation mask for the input image, which represents pixel-wise semantic distinctions for individual instances of objects. Specifically, each pixel is assigned a pair of an (unique) instance identifier (ID) $i \in \mathcal{I}$ and the corresponding object (or semantic) label $\ell_{i} \in  \mathcal{L}$, where $\mathcal{L}$ and $\mathcal{I}$ denote the sets of object labels (e.g., for indicating person, car, etc.) and instance IDs (e.g., for indicating multiple people, different cars, etc.), respectively.

\textbf{Instance-Aware Masking.}
Another innovation in our framework is to incorporate an instance-aware masking strategy into TTA for MDE.
Note that as described mathematically in \eqref{eq_1}, the key to MDE is to capture and extract the depth information from variations or changes in different frames, implying that dynamic objects (e.g., cars, people, animals, moving machinery, etc.) contain useful information for MDE.
Unfortunately, however, most real-world monocular images include numerous static objects including the background elements, which even occupy overwhelmingly large spatial portions of the images. 
Obviously, the dominance of such static components significantly hinders the extraction of useful depth information and acquisition of accurate depth estimates (i.e., severely limits the MDE performance) during TTA.
To address this critical issue, in our instance-aware masking strategy, we effectively suppress the impacts of the static objects on MDE during TTA by leveraging the so-called instance-wise masks.
Meanwhile, it has been widely recognized that among various visual cues, identifying the information on objects' or instances' shapes plays the most important role in many computer vision tasks including MDE {\color{black}\cite{landau1988importance}}. As a well-known example, Transformer-based networks tend to concentrate more on the shape information, which in turn contributes to their superior generalization behaviors to convolutional neural networks (CNNs) that tend to concentrate more on the texture information {\color{black}\cite{ballester2016performance, paul2022vision}}.
Inspired by these useful insights, in our instance-aware masking strategy, we produce the instance-wise masks from the panoptic segmentation mask to effectively extract the useful information on each instance's shape and we further utilize these masks to improve depth predictions.
We create the instance-wise masks only for the dynamic objects since the static ones are less informative for MDE during TTA as mentioned before.


Our instance-aware masking strategy proceeds as follows. 
Let $\mathcal{O} \subset \mathcal{L}$ denote the set of labels for the dynamic objects and $\mathcal{N} \subset \mathcal{I}$ denote the set of the corresponding instance IDs (i.e., the set of $i$'s corresponding to $\ell_{i} \in \mathcal{O}$) with cardinality $N$.
Then $N$ different segmentation masks for $N$ distinct dynamic instances in $\mathcal{N}$ are produced, which are referred to as the instance-wise masks in our framework and are denoted by $M_{j} \in  \mathbb{R}^{H \times W}$, $j =1,\cdots, N$, in Fig. \ref{fig:arch}.
Specifically, for such a segmentation mask corresponding to a dynamic instance $i \in \mathcal{N}$,
the intensity of each pixel $(x,y)$ is computed as
\begin{align}
\small
\label{IA_mask}
    g_{i}(x,y) = \begin{cases}
        i,  &   {\rm if~}  (x,y) \in  \mathcal{R}_{i} \\
        0 , &   {\rm otherwise}
    \end{cases}
\end{align}
where $\mathcal{R}_{i}$ is the segment assigned to the instance-object pair $(i, \ell_{i}) \in \mathcal{N} \times \mathcal{O}$.
Note that by \eqref{IA_mask}, the shape information about all the instances irrelevant to $i \in \mathcal{N}$ (including all static objects/instances) can be removed while maintaining the shape information of a dynamic instance $i \in \mathcal{N}$ of interest.
To this end and for more clear and delicate distinctions between the selected and remaining instances, in PITTA, the depth map is refined by applying the instance-wise masks as 
\begin{align}
\label{ins_mask}
    \tilde{D}_{j} & = D \odot M_{j}, \quad j=1,\cdots, N,
\end{align}
where $\odot$ stands for the element-wise multiplication.
These masked depth maps will be further used in the loss function design to properly refine the depth estimates.

\textbf{Edge Extraction.}
We also discern that the boundaries (or edges) of both dynamic and static objects in images are highly useful and informative for MDE during TTA as they can offer additional structural or geometric cues that can further enhance or refine the depth predictions, for instance, by properly guiding the alignment of predicted depth discontinuities with true object boundaries {\color{black}\cite{godard2017unsupervised}}.
Inspired by this, in PITTA, we introduce a weighted edge extraction approach based on the Laplacian approximation. 
In our edge extraction methodology, we construct the edge maps for the input image and the predicted depth map, denoted by $\partial I \in  \mathbb{R}^{H \times W}$ and $\partial D \in  \mathbb{R}^{H \times W}$ in Fig. \ref{fig:arch}, respectively, via 
\begin{align}
\label{I_edge}
    \partial I (x,y)  =  U(x,y) \times \big| &  \bar{I} (  \min  \{ x+1, H-1  \} , y  )   +   \bar{I} (  \max  \{ x-1, 0 \} , y  )  \nonumber\\
    & +   \bar{I} (  x, \min  \{ y+1, W-1  \}  )   +   \bar{I} (  x, \max  \{ y-1, 0 \}   )  -  4 \bar{I}(x,y)    \big|  ,  \\
    \label{D_edge}
    \partial D (x,y)  =  V(x,y) \times \big| &  D (  \min  \{ x+1, H-1  \} , y  )   +   D (  \max  \{ x-1, 0 \} , y  ) )  \nonumber\\
    &  +   D (  x, \min  \{ y+1, W-1  \}   +   D (  x, \max  \{ y-1, 0 \}   )  -  4 D(x,y)    \big|  , 
\end{align}
for $x \in  \mathcal{H} \triangleq \{  0, 1, \cdots, H-1 \}$ and $y \in  \mathcal{W} \triangleq \{  0, 1, \cdots, W-1 \}$, where $ U(x,y)$ and $ V(x,y)$ are nonnegative weights.
Also, $\bar{I} (x,y) =  \frac{1}{C} \sum_{z \in \mathcal{C}} I(x,y,z)$ for $ x \in \mathcal{H}$ and $ y \in \mathcal{W}$, where $\mathcal{C} \triangleq \{  0, 1, \cdots, C-1 \}$.
The extracted edge maps in \eqref{I_edge} and \eqref{D_edge} will be exploited in the loss function design for more precise depth estimation and refinement.
\subsection{TTA Methodology}

\textbf{Loss Functions.} PITTA adopts the two loss functions for adaptation of the MDE network: one is the depth-refining loss $L_{\rm d}$ and the other is the edge-guided loss $L_{\rm e}$. 
In the depth-refining loss, we leverage the median filtering technique to denoise the masked depth maps $\{  \tilde{D}_{j}  \}_{j=1}^{N}$ while retaining edge details as follows:

\begin{align}
\label{pseudo_label}
    \tilde{D}_j'(x,y)  =   {\rm median} \Big\{   \tilde{D}_{j}(p,q)  :  ~ &  p \in  \big[  \max \{x-r, 0 \}, \min\{x+r, H-1 \} \big] ,  \nonumber\\
    &  q \in  \big[ \max\{ y-r ,0  \},  \min \{ y+r, W-1 \} \big]   \Big\}, ~ j=1,\cdots, N,
\end{align}

for $x \in  \mathcal{H} $ and $y \in  \mathcal{W} $, where $r = \lfloor s/2 \rfloor$ and $s$ denotes the window size. 
The denoised versions $\{  \tilde{D}_{j}'  \}_{j=1}^{N}$ are used as pseudo labels for the depth estimates. The depth-refining loss is formulated as the squared Euclidean distance between the masked depth maps and their denoised counterparts as follows:
\begin{align}
\small
L_{\rm d} &=  \frac{1}{N} \sum_{j=1}^{N}  \Big\|  {\rm vec} ( \tilde{D}_{j} ) - {\rm vec} ( \tilde{D}_{j}' )   \Big\|_1  =  \frac{1}{N} \sum_{j=1}^{N} \sum_{(x,y) \in \mathcal{H} \times \mathcal{W}} \Big| D_j (x,y) - D_j'(x,y) \Big| 
\label{eq:depth}
\end{align}
where ${\rm vec} ( \cdot )$ is the vectorization operator.
Meanwhile, to compensate for potential misalignment or discrepancy between the edge maps $\partial I$ and $\partial D$--corresponding to the input image and the predicted depth map, respectively--with additionally taken into consideration their inherent sparse nature, we also introduce the edge-guided loss, which is defined as the Manhattan distance between the two edge maps $\partial I$ and $\partial D$ as follows:
\begin{align}
\small
L_{\rm e} &= \big\|  {\rm vec} ( \partial I ) - {\rm vec} ( \partial D )   \big\|_{1}  =   \sum_{(x,y) \in \mathcal{H} \times \mathcal{W}}  \Big| \partial I (x,y) - \partial D (x,y) \Big|.
\label{eq:edge}
\end{align}
The above two loss functions $L_{\rm d}$ and $L_{\rm e}$ are then integrated into one overall loss function via a weighted combination:
\begin{equation}
    L = L_{\rm d} + \lambda L_{\rm e}
    \label{eq:total}
\end{equation}
where $\lambda \geq 0$ controls the tradeoff between the two loss functions $L_{\rm d}$ and $L_{\rm e}$.

\textbf{Adaptation of MDE Network.} To effectively cope with the catastrophic forgetting issue while maintaining the generalization ability during TTA, in our framework, we suggest to adapt only some parts of the MDE network while fixing the remaining parts. It should be noted that our approach of selectively adapting the MDE network is clearly distinct from and more general than prior approaches in {\color{black}\cite{kuznietsov2021comoda, li2023test}} adapting the whole part of the MDE network. 
Specifically, let $\theta$ denote a set of selected parameters to be adapted in the MDE network.
PITTA continually adapts $\theta$ by minimizing the overall loss function $L$ in \eqref{eq:total} as follows:
\begin{align}
\label{theta_upd}
    \theta \leftarrow  \theta - \alpha \nabla_{\theta} L
\end{align}
where $\alpha \geq 0$ is the learning rate and $\nabla_{\theta}$ denotes the gradient operator with respect to $\theta$.

\section{Experiments}

\subsection{Experimental Setup}

We examine the performance of our framework and compare it with other recently developed competing TTA techniques including {\color{black}CoMoDa \cite{kuznietsov2021comoda}, Ada-Depth \cite{li2023test}, ActMAD \cite{mirza2023actmad}, and CoTTA \cite{wang2022continual}}. 
For fair comparisons, these competing methods are adjusted to accommodate our depth-refining loss with instance-aware masking strategy by using the panoptic segmentation network. 
We also report the performance of {\color{black}DepthAnythingV2 \cite{yang2024depth}}---a synthetic data-based full training approach---as benchmark for our proposed method.
In both our framework and other competing approaches, the MDE network of MonoDepth2 \cite{godard2019digging} pretrained on the KITTI dataset \cite{geiger2012we}---a standard benchmark dataset for MDE---is adopted as backbone and it is adapted to two other different datasets during TTA: (i) DrivingStereo dataset \cite{yang2019drivingstereo} and (ii) Waymo dataset \cite{sun2020scalability}.
In PITTA, we also employ Mask2Former \cite{cheng2022masked} with Swin-Tiny \cite{liu2021swin} backbone as the panoptic segmentation network.
For test cases with MonoDepth2, the adapted parameters in PITTA are chosen as the parameters of batch normalization (BN) layers in the encoder of the MDE network, i.e., $\theta = \gamma \cup \beta$, where $\gamma$ and $\beta$ represent the sets of scale and shift parameters in the BN layers of the encoder, respectively.
In the experimental evaluations, we measure the following popular performance metrics for MDE {\color{black}\cite{eigen2014depth}}: absolute relative error (AbsRel), square relative error (SqRel), root mean square error (RMSE), log-scale RMSE (RMSElog), and threshold accuracies ($\delta <  1.25$, $\delta <  1.25^{2}$, and $\delta <  1.25^{3}$). 
More details about these performance metrics are provided in {\color{black}Appendix D}.
Further details on the experiment setting and dataset preprocessing can be found in {\color{black}Appendices E and F}, respectively. 

\subsection{Results and Discussion}


Table \ref{table:main_res} shows results of our framework and other approaches on the adaptation to the DrivingStereo dataset for test cases under diverse weather conditions---foggy, rainy, and mixture of all weather conditions. 
It could be observed that for each test case, PITTA surpasses the other TTA methods in all performance metrics with remarkable improvements.
This thus proves the superiority, effectiveness, and robustness of our TTA strategy for MDE based on pose-agnostic adaptation and instance-aware masking.
Table \ref{table:Waymo} also presents results of the various methods on the adaptation to the Waymo Dataset---known as a more challenging dataset than the DrivingStereo dataset {\color{black}\cite{yang2019drivingstereo}}.
It shows that our framework attains the best performance in all test cases, thereby validating its universality and resilient adaptability across various domains.
In Table \ref{table:comp_depth}, the performance of our framework is further demonstrated when it is applied to adaptations of other MDE networks---{\color{black}SGDepth \cite{guizilini2020semantically}, HR-Depth \cite{lyu2021hr}, Lite-Mono \cite{zhang2023lite}, and MonoViT \cite{zhao2022monovit}}---that are pretrained on the KITTI dataset to the DrivingStereo dataset. 
More results of our proposed method for these MDE networks can be found in {\color{black}Appendix G.1}. Extended results for Mask2Former---the panoptic segmentation network adopted in PITTA---with different backbones are also available in {\color{black}Appendix G.2}.
It shows that PITTA can effectively adapt different types of MDE networks with performance improvements, where adapting SGDepth is observed to be the most effective by exhibiting the most significant improvements due to the network-architectural similarity to our framework.
This manifests versatility of our proposed adaptation strategy.

\begin{table*}[!t]
    \centering
    \caption{\small Results of our proposed method and other competing methods applied to adaptation of MDE network of MonoDepth2 pretrained on KITTI dataset to DrivingStereo dataset. In each test case, numbers in \textbf{bold} and \underline{underline} indicate the best performance and the second best performance, respectively. Also, ``All'' indicates a test case concatenating cloudy, foggy, rainy, and sunny weather conditions in turn.}
    \label{table:main_res}
    \begin{adjustbox}{max width=0.8\textwidth}
    \begin{tabular}{ccccccccc}
    \toprule
       Weather & Method & AbsRel ↓ & SqRel ↓ & RMSE ↓ & RMSElog ↓ & $\delta<1.25$ ↑ & $\delta<1.25^2$ ↑ & $\delta<1.25^3$ ↑\\
       \midrule
        \multirow{7}{*}{Foggy}
            & No adaptation       & \underline{0.143} & \underline{1.952} & \underline{9.817} & \underline{0.218} & \underline{0.812} &   \underline{0.937} & \underline{0.974} \\
            & CoMoDA       &   0.566  &  10.914  &  21.392  &   0.657  &   0.249  &   0.467  &   0.660  \\
            & CoTTA        &   0.175  &   2.344  &   9.899  &   0.243  &   0.750  &   0.920  &   0.969  \\
            & ActMAD       &   0.176  &   2.403  &  11.157  &   0.262  &   0.730  &   0.902  &   0.957 \\
            & Ada-Depth    &   0.208  &   3.306  &   9.884  &   0.227  &   0.719  &   0.884  &   0.948  \\
            & DepthAnythingV2 &   0.164    &   2.741   &   12.891  &    0.278     &   0.736    &    0.880     &     0.944 \\
            \rowcolor{lightgray!50}
            & \textbf{PITTA (ours)} & \textbf{0.127} & \textbf{1.579} & \textbf{8.600} & \textbf{0.195} & \textbf{0.840} & \textbf{0.951} & \textbf{0.980} \\
        \midrule
        \multirow{7}{*}{Rainy} 
            & No adaptation       & 0.245 & 3.641 & 12.282 & 0.310 & 0.600 & 0.852 & 0.945 \\
            & CoMoDA       &   0.701  &  18.848  &  22.759  &   0.727  &   0.303  &   0.497  &   0.653  \\
            & CoTTA         &   0.225  &   3.127  &  11.057  &   0.286  &   0.638  &   0.875  &   0.960  \\
            & ActMAD       & 0.256 & 3.631 &12.399  & 0.328 & 0.590 & 0.818 & 0.936\\
            & Ada-Depth    &   0.322  &   5.623  &  10.782  &   0.387  &   0.572  &   0.794  &   0.897  \\
            & DepthAnythingV2 &   0.208    &   3.105   &   12.480  &    0.293     &   0.657    &    0.860     &     0.946 \\
            \rowcolor{lightgray!50}
            & \textbf{PITTA (ours)} & \textbf{0.195} & \textbf{2.532} & \textbf{10.315} & \textbf{0.254} & \textbf{0.685} & \textbf{0.905} & \textbf{0.973} \\
        \midrule
        \multirow{7}{*}{All} 
            & No adaptation       & 0.181  &   2.446  &   9.536  &   0.247  &   \underline{0.749}  &   0.913  &   0.965\\
            &  CoMoDA        &  0.577 & 11.736 & 20.289 &  0.673 &  0.277 &  0.497 &  0.678 \\    
            & CoTTA          &  0.200 &  2.619 &  9.657 &  0.265 &  0.710 &  0.899 &  0.961 \\  
            & ActMAD       & 0.172  &   2.190  &   9.543  &   0.244  &   \textbf{0.758}  &   0.913  &   0.967\\
            & Ada-Depth      &  0.272 &  4.417 &  10.512 &  0.339 &  0.624 &  0.833 &  0.921 \\  
            & DetphAnythingV2 &   0.187    &   2.669   &   11.883  &    0.289     &   0.705    &    0.880     &     0.946 \\
            \rowcolor{lightgray!50}
            & \textbf{PITTA (ours)} & \textbf{0.171}  &   \textbf{2.016}   &   \textbf{9.200}    &  \textbf{0.241}   &   \textbf{0.758}   &   \textbf{0.918}   &   \textbf{0.968} \\
            \bottomrule
    \end{tabular}
    \end{adjustbox}
\end{table*}

\begin{table*}[!t]
    \centering
    \caption{\small Results of our proposed method and other competing methods applied to adaptation of MDE network of MonoDepth2 pretrained on KITTI dataset to Waymo dataset. Other conventions are the same as in Table \ref{table:main_res}.}
    \begin{adjustbox}{max width=0.8\textwidth}
    \begin{tabular}{cccccccc}
    \toprule
         Method &  AbsRel ↓ & SqRel ↓ & RMSE ↓ & RMSElog ↓ & $\delta<1.25$ ↑ & $\delta<1.25^2$ ↑ & $\delta<1.25^3$ ↑\\
         \midrule
         No adaptation &  0.219  &   2.912  &   9.060  &   0.284  &   0.658  &   0.879  &   0.955\\
         CoMoDA & 0.543  &   8.098  &  14.309  &   0.597  &   0.316  &   0.517  &   0.693\\
         CoTTA    &  0.273 &  4.239 &  10.347 &  0.331 &  0.584 &  0.835 &  0.933 \\ 
         ActMAD & 0.230  &   2.559  &   8.866  &   0.292  &   0.621  &   0.873  & 0.960 \\
         Ada-Depth     &  0.315 &  3.462 &  9.809 &  0.377 &  0.451 &  0.771 &  0.921 \\
         DetphAnythingV2 &   0.221    &   2.431   &   9.473  &    0.309     &   0.614    &    0.849     &     0.943 \\
         \rowcolor{lightgray!50}
         \textbf{PITTA (ours)} &   \textbf{0.199}     &    \textbf{2.097}   &    \textbf{8.206}   &     \textbf{0.266}    &   \textbf{0.670}    &     \textbf{0.896}   &   \textbf{0.967}  \\
         \bottomrule
    \end{tabular}
    \end{adjustbox}
    \label{table:Waymo}
\end{table*}

\begin{table*}[!t]
    \centering
    \caption{\small Results of our proposed method applied to adaptations of various MDE networks pretrained on KITTI dataset to DrivingStereo dataset. Other conventions are the same as in Table \ref{table:main_res}.}
    \begin{adjustbox}{max width=0.8\textwidth}
    \begin{tabular}{clccccccc}
    \toprule
        MDE Net & Weather &  AbsRel ↓ & SqRel ↓ & RMSE ↓ & RMSElog ↓ & $\delta<1.25$ ↑ & $\delta<1.25^2$ ↑ & $\delta<1.25^3$ ↑\\
       \midrule
        \multirow{3}{*}{SGDepth} 
        & No adaptation  & 0.207 & 2.527 & 9.738 & 0.290 & 0.725 & 0.893 & 0.948\\
        & Sunny  & 0.174 & 1.955 & 8.334 & 0.245 & 0.781 & 0.922 & 0.965 \\ 
        & All &  0.170      &   1.809   &   8.305  &    0.244     &    0.784    &     0.922     &     0.964 \\
        \midrule
        \multirow{3}{*}{HR-Depth}
        & No adaptation & 0.164 & 1.694 & 7.617 & 0.225 & 0.783 & 0.938 & 0.977\\
        & Sunny &   0.171     &   1.711   &    7.970  &     0.240    &    0.787    &     0.924     &   0.964  \\
        & All &  0.169 & 1.865 & 9.022 & 0.255 & 0.776 & 0.913 & 0.959 \\
        \midrule
        \multirow{3}{*}{MonoViT}
        & No adaptation &  0.150 & 1.609 & 7.648 & 0.211 & 0.815 & 0.943 & 0.979\\
        & Sunny &   0.141     &   1.425   &    7.528  &     0.207    &    0.826    &     0.948     &   0.979  \\
        & All  &  0.150      &   1.550   &   7.609  &    0.213     &    0.814    &     0.938     &     0.977 \\  
        \midrule
        \multirow{3}{*}{Lite-Mono}
        & No adaptation &  0.184 & 2.166 & 8.383 & 0.248 & 0.766 & 0.918 & 0.965\\
        & Sunny  &   0.171     &   1.844   &    7.970  &     0.240    &    0.787    &     0.924     &   0.964  \\
        & All  &  0.173      &   2.079   &   9.406  &    0.243     &    0.750    &     0.915     &     0.968    \\
        \midrule
        \multirow{3}{*}{DetphAnythingV2}
        & No adaptation &   0.194    &   2.473   &   11.197  &    0.300     &   0.707    &    0.884     &     0.944 \\ 
        & Sunny  &   0.194    &   2.498   &   11.322  &    0.302     &   0.708    &    0.882     &     0.943 \\
        & All  &   0.198    &   3.001   &   12.593  &    0.305     &   0.684    &    0.857     &     0.935    \\
    \bottomrule
    \end{tabular}
    \end{adjustbox}
    \label{table:comp_depth}
\end{table*}

\subsection{Ablation Study}


\textbf{Effect of Loss Functions.} 
As an ablation study on our framework, the tradeoff between the depth-refining loss $L_{\rm d}$ and edge-guided loss $L_{\rm e}$ is examined in Table \ref{tab:ablation_lambda} in terms of the AbsRel metric and threshold accuracy with $\delta < 1.25$ for the case of adaptation to the Waymo dataset. The results demonstrate that the performance of PITTA is sensitive to the choice of the tradeoff parameter $\lambda$. Specifically, as $\lambda$ increases, the performance initially improves (up to $\lambda=0.2$) and then deteriorates, indicating that there exists an optimal operating point with respect to $\lambda$. This in turn suggests that in practical design, more careful consideration is needed in the selection of $\lambda$ as PITTA will be most effective with modest values of $\lambda$. 
More results on this ablation study for other metrics are available in {\color{black}Appendix G.3}.

\begin{table}[!t]
    \centering
    \caption{\small Ablation study on $\lambda$ regarding tradeoff between two loss functions $L_{\rm d}$ and $L_{\rm e}$ in terms of AbsRel and threshold accuracy $\delta < 1.25$ for adaptation to Waymo dataset. Other conventions are the same as in Table \ref{table:main_res}.}
    \label{tab:ablation_lambda}
    \begin{adjustbox}{max width=0.8\textwidth}
    \begin{tabular}{ccccccccccccc}
    \toprule
    $\lambda$   &   0   &   0.1    &    0.2    &    0.3     &   0.4     &   0.5     &   0.6     &   0.7     &  0.8     &   0.9     &   1.0     &       $\infty$    \\
    \midrule
    AbsRel ↓ &   0.236   &   0.203   &   1.999   &    0.200   &  0.202   &   0.204   &  0.205   &   0.206   &   0.207   &   0.208   &   0.208   &   0.218   \\
    $\delta < 1.25$ ↑ &   0.618   &   0.644   &   0.670   &     0.668   &   0.663   &   0.657   &   0.654   &   0.652   &   0.649   &   0.647   &   0.647   &   0.627   \\
    \bottomrule
    \end{tabular}
    \end{adjustbox}
\end{table}

\begin{table}[!t]
    \centering
    \caption{\small Ablation study on the number of adapted parameters in the MDE network of MonoDepth2 in terms of AbsRel and threshold accuracy $\delta < 1.25$ for adaptation to Waymo dataset. ``CNN $a$\% + BN $b$\%'' indicates that only the parameters of last $a$\% of CNN layers and last $b$\% of BN layers in the encoder of the MDE network are adapted. Other conventions are the same as in Table \ref{table:main_res}.}
    \label{tab:ablation_param}
    \begin{adjustbox}{max width=0.9\textwidth}
    \begin{tabular}{cccccc}
    \toprule
       fraction &   CNN 0\% + BN 50\%  &   CNN 0\% + BN 100\%   &  CNN 20\% + BN 80\%  &  CNN 40\% + BN 60\%  & CNN 60\% + BN 40\%   \\
       \midrule
       AbsRel ↓ &   0.214   &   0.199   &   0.828   &   0.602   &   0.770   \\
       $\delta < 1.25$ ↑ &  0.636   &   0.670   &   0.241   &   0.248   &   0.284   \\
    \bottomrule
    \end{tabular}
    \end{adjustbox}
\end{table}

\textbf{Impact of Number of Adapted Parameters.} 
In Table \ref{tab:ablation_param}, we conduct another ablation study on the number of adapted parameters of the encoder (i.e., cardinality of $\theta$) in the MDE network of MonoDepth2 based on AbsRel metric and threshold accuracy with $\delta < 1.25$ for the case of adaptation to the Waymo dataset. 
It confirms that the performance crucially depends on the number of adapted parameters because of a conflict between the catastrophic forgetting and adapting to new domains. 
In practical usage scenarios, therefore, the degree or amount of adaptation on the MDE network should be judiciously determined. 
For the MDE network of MonoDepth2, adapting only the parameters in the BN layers of the encoder (as described previously in Section 4.1) turned out to excel in almost all performance metrics in our experimental setup.
Extended results of this ablation study for other MDE metrics and other MDE networks are available in {\color{black}Appendix G.4}.

\section{Conclusions}

In this work, we introduced PITTA, an innovative and novel TTA framework for MDE, markedly outperforming the prior SOTA methods in online adaptation of pretrained MDE network without accesses to source datasets and camera pose information.
By leveraging instance-wise masks for dynamic objects in a novel and elegant manner, our method substantially enhanced the adaptability of MDE network during TTA. On top of this, we also devised effective edge extraction methodology and loss function design to further improve the MDE network's adaptability.
With these strategies, our framework in turn exhibited superior performance and better effectiveness over other competing methods, as substantiated by thorough and extensive experimental validation on Driving Stereo and Waymo datasets, thereby underlining its good potential and wide applicability in TTA for MDE.
Ablation studies further confirm efficacy and superiority of our model design choices. 
The scope and applicability of our TTA framework may be confined to MDE tasks, and thus, it is deserved to study the extendability and generality of our framework to other tasks as further works.

\bibliographystyle{unsrt}
\bibliography{references}

\end{document}